\documentclass[runningheads]{llncs}
\usepackage{graphicx}
\usepackage [misc]{ifsym}
\usepackage{bbding}
\usepackage{placeins}

\usepackage{hhline}
\begin{document}
\title{BERT-CoQAC: BERT-based Conversational Question Answering in Context}
%
%
\author{Munazza Zaib\inst{1} \thanks{Corresponding author.}  \and
Dai Hoang Tran\inst{1} \and Subhash Sagar \inst{1} \and Adnan Mahmood \inst{1} \and Wei E. Zhang \inst{1,2} \and
Quan Z. Sheng \inst{1}}
\authorrunning{M. Zaib et al.}
%
\institute{Department of Computing, Macquarie University, Sydney, NSW 2109, Australia \email{munazza-zaib.ghori@hdr.mq.edu.au} \and School of Computer Science, The University of Adelaide, Adelaide, Australia} 

\maketitle              
\begin{abstract}
As one promising way to inquire about any particular information through a dialog with the bot, question answering dialog systems have gained increasing research interests recently. Designing interactive QA systems has always been a challenging task in natural language processing and used as a benchmark to evaluate machine’s ability of natural language understanding. However, such systems often struggle when the question answering is carried out in multiple turns by the users to seek more information based on what they have already learned, thus, giving rise to another complicated form called \textit{Conversational Question Answering (CQA)}. CQA systems are often criticized for not understanding or utilizing the previous context of the conversation when answering the questions. To address the research gap, in this paper, we explore how to integrate the conversational history into the neural machine comprehension system. On one hand, we introduce a framework based on publicly available pre-trained language model called BERT for incorporating history turns into the system. On the other hand, we propose a history selection mechanism that selects the turns that are relevant and contributes the most to answer the current question. Experimentation results revealed that our framework is comparable in performance with the state-of-the-art models on the QuAC\footnote{http://quac.ai/} leader board. We also conduct a number of experiments to show the side effects of using entire context information which brings unnecessary information and noise signals resulting in a decline in the model's performance.

\end{abstract}

\section{Introduction}
\label{intro}
The field of conversational AI can be divided in to three categories namely, {\em goal-oriented dialogue systems}, {\em chat-oriented dialogue systems}, and {\em question answering (QA) dialogue systems}.  
The former two have been very researched upon topics, resulting in a number of successful dialogue agents such as Amazon Alexa, Apple Siri and Microsoft Cortana. However, QA dialogue systems are fairly new and still require extensive research. To facilitate the growth of this category, many question answering challenges were proposed \cite{rajpurkar-etal-2016-squad,joshi-etal-2017-triviaqa,kocisky-etal-2018-narrativeqa} which later gave rise to the field of {\em Conversational Question Answering} (CQA). CQA has introduced a new dimension of dialogue systems that combines the elements of both chit-chat and question answering. Conversational QA is a ``user ask, system respond'' kind of setting  where a user starts the conversation with a particular question or information need and the system searches its database to find an appropriate solution of that query. This could turn into a multi-turn conversation if 
the user needs to have more detailed information about the topic. The ability to take into account previous utterances is key to building interactive QA dialogue systems that can keep conversations active and useful. Yet, modeling conversation history in an effective way is still an open challenge in such systems.

Existing approaches have tried to address the problem of history conversation modeling
by prepending history questions and answers to the current question and source passage \cite{DBLP:journals/tacl/ReddyCM19}. Though this seems to be a simple method to improve the answer's accuracy, in reality it fails to do so. Another approach used complex Graph Neural Networks \cite{DBLP:journals/corr/abs-1908-00059} to deal with this issue. One recent work \cite{DBLP:conf/sigir/Qu0QCZI19} introduced the use of history answer embeddings but they 
did not consider using relevant context rather than used entire history turns to find the answer span. Also, they did not draw the complete picture of the context to the model by eliminating history questions. 
Table~\ref{tab:1}
shows a chunk of dialogue extracted from the QuAC \cite{choi-etal-2018-quac} dataset. In order to answer the query Q2, we expect the system to have the knowledge of Q1 and A1, so that it can easily decipher the ``he'' entity in Q2. This shows that modeling complete history is necessary when designing an effective conversational QA system.
\FloatBarrier
\begin{table}[!h]\centering
{\renewcommand{\arraystretch}{1.15}}
\begin{tabular}{p{1.5cm}p{1.5cm}p{5.5cm}}
      \hline
    \hline
     \multicolumn{3}{c}{Topic: Formative years and life's calling} \\
     \hline
    \textbf{ID} & \textbf{Role} & \textbf{Conversation} \\
    \hline
    Q1 & Usr & When was Kurien born? \\
    A1 & Sys & He was born on 26 November, 1921 \\\hline
    Q2 & Usr & Where was he born? \\
    A2 & Sys & Calicut, Madras Presidency (now Kozhikode, Kerala) in a Syrian Christian family. \\
    \hline
    \hline
    \end{tabular}
    \caption{A chunk of a dialogue from QuAC dataset.}
    \label{tab:1}
\end{table}
\FloatBarrier
To address this shortcoming, in this paper, we emphasize the following research questions:

 \paragraph{\textbf{Research Question 1:}} \textit{How can we utilize the context in an efficient way?} To answer this, we propose an effective framework, called \textbf{B}idirectional \textbf{E}ncoder \textbf{R}epresentations from \textbf{T}ransformers based \textbf{C}onversational \textbf{Q}uestion \textbf{A}nswering in \textbf{C}ontext (BERT-CoQAC), that uses a mechanism to only extract the context that is relevant to the current question by calculating the similarity score between the two. By doing so, our model would be able to generate better contextualized representation of the words, thus resulting in the better answer spans. Apart from this, we also propose the use of history questions to support and improve the effect of history answer embeddings when looking for an answer span. 
 
\paragraph{\textbf{Research Question 2:}} \textit{What is the effect of incorporating the entire context into conversational question answering system?} To answer this, we perform extensive experiments and finds out that using the entire context results in decreasing the model's performance due to the presence of unnecessary information in the provided input.

Thus, the contributions of our work are summarized as the following:
i) we introduce a new method of incorporating selected history questions along with answer embeddings to model the complete conversation history; ii) we present that noise in context utterances could result in decline in model's performance; 
 iii) the experimental results shows that our method achieves better performance in accuracy than the other different state-of-the-art published models.
 \section{Related Work}
The concept of BERT-CoQAC is similar to machine comprehension and can be termed as conversational machine comprehension (CMC). The difference between MC and CMC is that questions in MC are independent of each other whereas questions in CMC form a series of questions that requires a proper modeling of the conversation history in order to comprehend the context of current question correctly. Different models 
and approaches have been proposed to handle the complete conversation. 
\cite{DBLP:conf/aaai/SerbanSBCP16} used hierarchical models, first capturing the meaning of individual utterances and then combining them as discourses. 
\cite{DBLP:conf/aaai/XingWWHZ18} extended this concept with the attention mechanism to attend to significant parts of the utterances on both word and utterance level, respectively. In another study \cite{DBLP:conf/acl/TianYMSFZ17}, a systematic comparison between hierarchical and non-hierarchical methods was conducted and the authors proposed a variant that weighs the context with respect to context-query relevance.

High quality conversational dataset such as QuAC \cite{choi-etal-2018-quac} and CoQA \cite{DBLP:journals/tacl/ReddyCM19} have provided the researchers a great source to work deeply in the field of CMC. In our work, we choose to work on QuAC because it encourages users to participate more in the information seeking dialogue. In this setting, the information seeker has the access only to the title of the paragraph and can pose free-form questions to learn about the hidden text of Wikipedia paragraph.

The baseline model for QuAC is based on single turn machine comprehension model known as BiDAF \cite{DBLP:conf/iclr/SeoKFH17} which was further extended to BiDAF++ w/x-ctx \cite{choi-etal-2018-quac} that marked the history answers in the source paragraph. BiDAF++ was further improved in BERT-HAE \cite{DBLP:conf/sigir/Qu0QCZI19}, where the concept of history answer embeddings was used. These embeddings were then concatenated with the given passage to identify the answer span. FlowQA \cite{DBLP:journals/corr/abs-1810-06683} introduced a mechanism to grasp the history of conversation by generating an intermediate representation of the previously carried out conversation. Later, Graphflow \cite{DBLP:journals/corr/abs-1908-00059} introduced a graph neural network (GNN) based model to construct the context-aware graph and capture the flow of the conversation.

The recent state-of-the-art advancements in pre-trained language models such as BERT, ELMo,  and GPT-2 have led to the rapid proliferation of research interests in the field of machine reading comprehension. These models can be utilized effectively in tasks with small datasets as the relationship between the words are already learnt by these models during the pre-training phase.
Out of all the pre-trained language models, BERT is known to produce state-of-the-art results in MRC tasks.

Just after a short period after its introduction, BERT has been widely used on different conversational datasets such as learning dialog context representations on MultiWOZ \footnote{http://dialogue.mi.eng.cam.ac.uk/index.php/corpus/} or evaluating the open-domain dialog system on DailyDialog \footnote{http://yanran.li/dailydialog.html} dataset.
On the QuAC leaderboard\footnote{https://quac.ai/}, various approaches \cite{DBLP:journals/corr/abs-1908-05117,DBLP:conf/sigir/Qu0QCZI19} use BERT as a base model for conversational question answering. The difference between aforementioned approaches and our model is that they either capture none or all of the previous conversational history turns whereas our proposed BERT-CoQAC finds relevant history turns in order to predict the accurate answer.
 \section{Methodology}
 \label{methods}
 In the following section, we present our model that addresses the two research questions described in Section~\ref{intro}.
 \subsection{Task Formulation}
Given a source paragraph \textit{P} and a question \textit{Q}, the task is to find an answer \textit{A} to the question provided the context and can be formulated as follows:

\textbf{Input:} The input of BERT-CoQAC consists of current question $Q_i$, given passage \textit{P}, history questions $Q_{i-1}, ..., Q_{i-k}$ and the embeddings of history answers $HE_{i-1}, ..., HE_{i-k}$,

\textbf{Output:} The answer $A_i$ identified using start span and end span generated by the model.

where \textit{i} and \textit{k} represents the indices of turn and the number of  dialogue history considered, respectively. 
\subsection{BERT-CoQAC}
\begin{figure}[h]
    \centering
\includegraphics[width=5in]{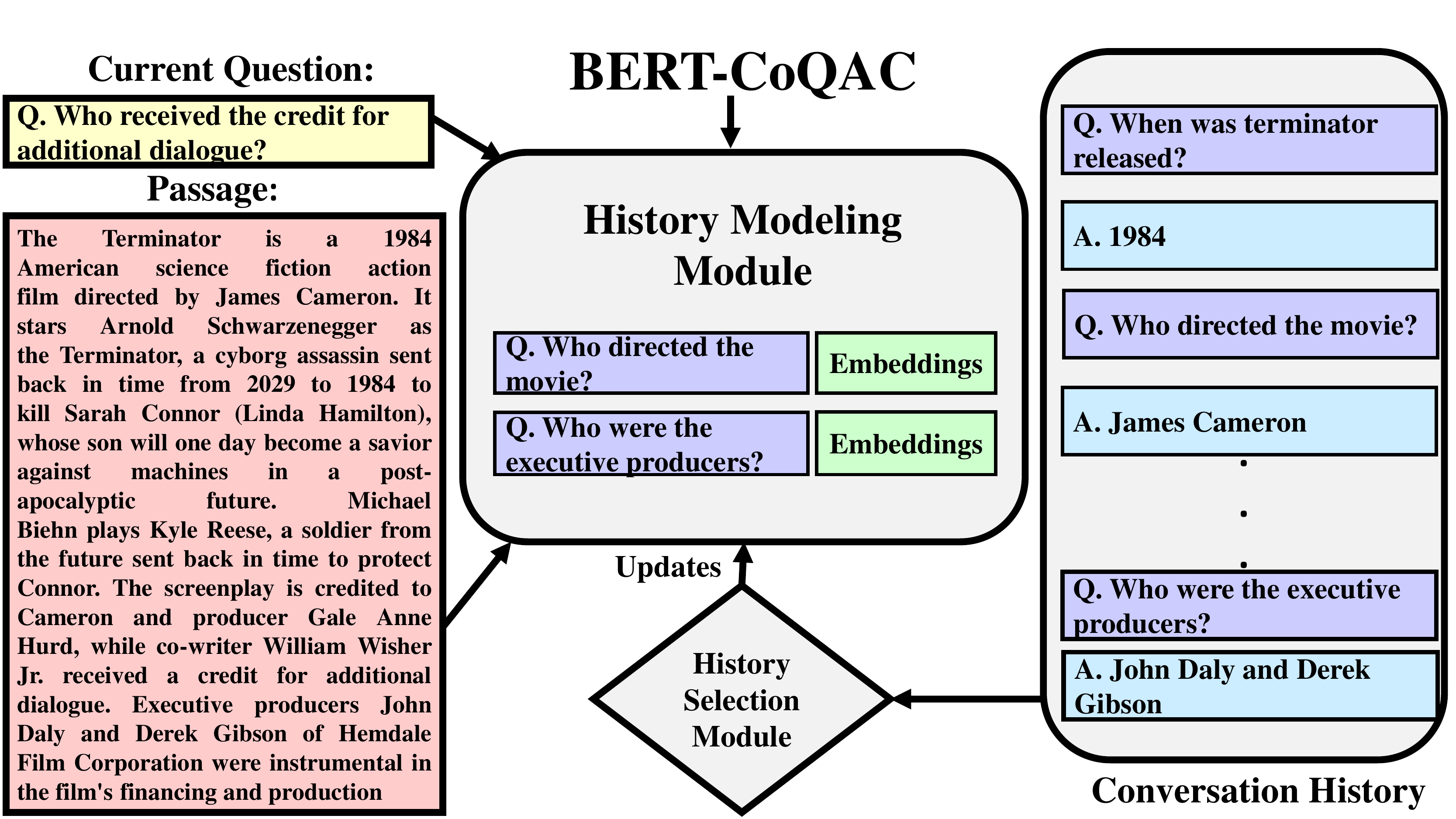}
    \caption{Modular representation of BERT-CoQAC. It shows the input formulation and the components of our model.}
    \label{modules}
\end{figure}

Fig.~\ref{modules} represents the overall framework of BERT-CoQAC, which 
consists of:
\subsubsection{History Selection Module}that calculates the context-query relevance score and selects the history turns that are expected to be more relevant to the question, and
\subsubsection{History Modeling Module} that takes the selected history turns along with the current question and passage, and transforms them into the format required by BERT.

\begin{figure}[h]
    \centering
    \includegraphics[width=12cm, height=6cm]{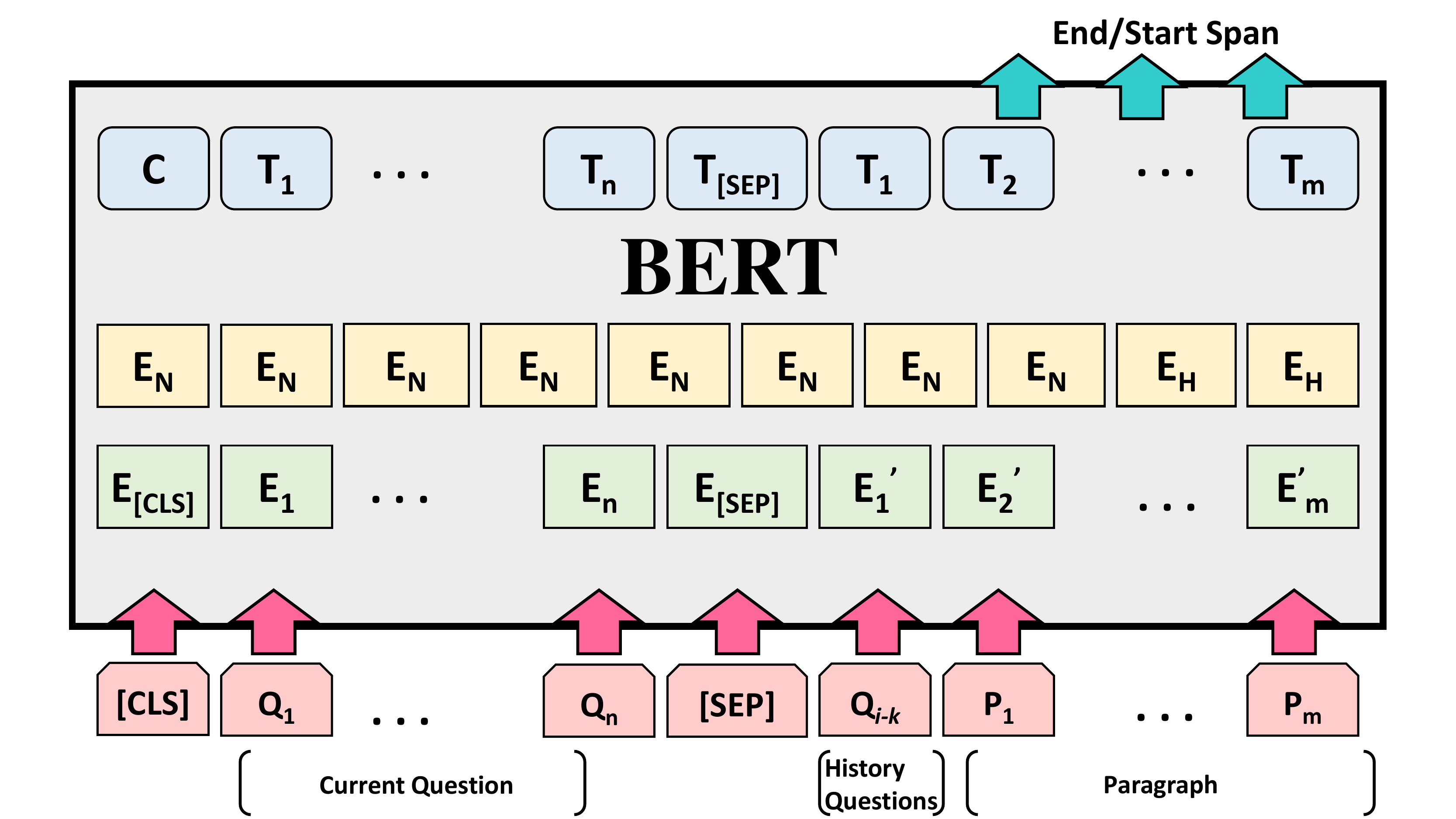}
    \caption{Architecture of BERT-CoQAC model. History questions are prepended with the passage and \textit{$E_N/E_H$} denotes whether the token is present in history or not.}
    \label{arch}
\end{figure}

 Inside the \textit{History Modeling} module, we leverage the strengths of pre-trained language model and adapt BERT to suit our task's requirements. Fig.~\ref{arch} illustrates our model's architecture that includes embeddings of history answers along with the history questions. Another factor worth noting here is that \textit{History Selection} module  only selects those history questions and answer embeddings that have cosine similarity score greater than the threshold value calculated using:
 \begin{equation}
      score_i=  \frac{h_i . q}{||h_i||.||q||}
 \end{equation}

 where $h_i$ denotes current history turn and \textit{q} represents the query. Finally, the scores are normalized using softmax function to get the probability distribution as shown in Equation~\ref{two}.
 \begin{equation}
      p_i= \frac{exp(score_i)}{\sum^n_{j_{=_0}} exp(score_i)}
      \label{two}
 \end{equation}

We performed different experiments and found out that turns having score greater than or equal to 0.5 contributes more in generating accurate answer span, thus, the threshold value is set to 0.5. Turns having value less than this will be filtered out. Our model feeds the conversation history to BERT in a natural way and generates two types of tokens for history answer embeddings. \textit{$E_H/E_N$} shows that whether the token is a part of history answer or not. These embeddings have influence on the information that the other tokens possess. The history questions are not part of the passage, so we cannot ``embed'' them directly to the input sequences. Instead, we prepend the historical questions in the same sequence as that of embeddings to improve the answer span. Let \textit{$T_i$} be the BERT-representation of the \textit{i\textsuperscript{th}} token and S be the start vector. The probability of this token being the start token is \(P_i =\frac{e\textsuperscript{$S.T_i$}}{\sum_k e\textsuperscript{S.$T_k$}}\). The probability of a token being the end token is computed likewise. The loss is the average of the cross entropy loss for the start and end positions. 

\section{Experimental Setup}
In this section, we describe the setup of our experiments for the evaluation of the proposed BERT-CoQAC model, including the dataset, model training, the comparison baseline methods, implementation details, and the evaluation metrics. 
\subsection{The Dataset}
The motivation behind QuAC\footnote{http://quac.ai/} (Question Answering in Context) comes from the idea of teacher-student setup where a student asks a series of questions about a topic to get in-depth information about it. The issue in this scenario is that most of the questions are context-dependant, can be abstract, and might not have any answer. It is up to the teacher to utilize and shift all the knowledge they have to provide the best answer. The seeker has access only to the heading of the given paragraph and the answers are provided by generating the start and end span in the paragraph. The training/validations sets have 11,000/1,000 questions across 14,000 dialogues. Every dialogue can have a maximum of 12 dialogue turns, which constitutes 11 history turns at most.
\subsection{Model Training}
\textit{($Q_i$, $Q_i^k$.P, {$HA_i^k$}, $A_i$)} denotes a single instance of training where \textit{$Q_i^k$.P} denotes the paragraph prepended with history questions in the same order as that of history answers, \textit{HA}. This instance is first transformed into example variation where each variation has only one history turn from the conversation history. A context-query relevance based history selection module then considers \textit{k} relevant history turns. A new instance, $(Q_i, Q_i^k.P, AE_i^k, A_i)'$, is formed by merging all the selected variations and used as an input to the BERT-CoQAC model. Since the length of the passages is greater than the maximum sequence length, therefore, we use the sliding window approach to split lengthy passages as suggested in the BERT \cite{DBLP:conf/naacl/DevlinCLT19} model.

\subsection{Comparison Systems}
A brief description of competing methods is as follows:
\begin{itemize} 
    \item \textbf{BiDAF++ \cite{DBLP:conf/iclr/SeoKFH17}:} BiDAF++ extends BiDAF by introducing contextualized embeddings and self-attention mechanism in the model.
    \item \textbf{BiDAF++ w/ 2-ctx \cite{choi-etal-2018-quac}:} It takes 2 history turns into account when extending BiDAF++. It also  concatenates the marker embeddings to passage embeddings and adds dialogue turn number into question embeddings.
    \item \textbf{FlowQA \cite{DBLP:journals/corr/abs-1810-06683}:} FlowQA introduces a mechanism that incorporates intermediate representations generated during the process of answering previous questions to make the model be able to grasp the latent semantics of the history.
    \item \textbf{BERT-HAE \cite{DBLP:conf/sigir/Qu0QCZI19}:} This model is built on pre-trained language model, BERT, to model history conversation using history answer embeddings.
\end{itemize}

Our proposed BERT-CoQAC framework is an improved model based on BERT-HAE with the introduction of history selection mechanism along with the history questions to improve the model's accuracy.
\subsection{Hyper-parameter Settings}
The model is implemented using Tensorflow and uses version v0.2 of QuAC dataset. We utilize the BERT-Base model (uncased) having maximum length of the sequence set to 384. Document stride is 128 and the maximum answer length is of 40. The batch size is set to 12. The turns to be incorporated are selected on the basis of their relevance to the current question.  The optimizer is Adam with a learning rate of 3e-5. The number of training epochs is 3. We use two NVIDIA Tesla P100 16GB GPUs.

\subsection{Evaluation Metrics}
For the evaluation purpose, we use not only the $F_1$ score but also the human equivalence score for questions (HEQ-Q) and dialogues (HEQ-D) \cite{choi-etal-2018-quac}. \textbf{HEQ-Q} represents the percentage of exceeding or matching the human performance on questions and \textbf{HEQ-D} represents the percentage of exceeding or matching the human performance on dialogues. 

\section{Evaluation Results}
Table~\ref{result} and Table~\ref{turns} shows the evaluation results of our model on the QuAC dataset. Our model outperforms the baseline methods and BERT-HAE model on all the three metrics i.e. $F_1$, HEQ-Q, and HEQ-D and answer our research questions as follows. 

\textbf{Research Question 1: } \textit{How  can  we  utilize  the  context  in  an  efficient  way? }

From Table~\ref{result}, we can make the following observations. i) Using conversation history has a significant effect when answering the current question. This holds true for both BiDAF++ and BERT-based methods. ii) Incorporating relevant history turns rather than the entire conversation has a significant effect when answering the current question. Our experiments confirms the hypothesis and outperforms the competing methods. iii) Apart from history answer embeddings, history questions also plays a significant part in improving answer's accuracy. The effect of including complete history turn (consisting of both question and answer) is presented more clearly in the next paragraph. iv) BERT-CoQAC with simple experiment setup performs just as good as FlowQA that uses convoluted mechanisms to model the history. v)The training time of our model is way more efficient than that of FlowQA and is slightly better than BERT-HAE as well which proves the efficiency of our proposed architecture. 

\textbf{Research Question 2:} \textit{What  is  the  effect  of  incorporating  the  entire  context into conversational question answering system?}  

We conducted extensive experiments without using context-query relevance mechanism and took maximum (i.e.11) number of turns into consideration. Table~\ref{turns} shows that selection of relevant history is essential to generate better answer spans. Utilizing entire context results in decreasing the model's performance. However, our model still performs better than BERT-HAE just by simply adding the history turns, comprising of both previous questions and answers, into the model. Our model provides the highest accuracy after introducing 5 history turns which is an improvement on BERT-HAE model that provides high accuracy after 6 turns which shows that introducing complete history is necessary for better answer accuracy rather than just using history answer embeddings. 
\begin{table}[h!]
\begin{center}
    \begin{tabular}{p{3.5cm}|p{2.5cm}|p{2.5cm}|p{2.5cm}|p{2.5cm}} \hline
    \hline
\multicolumn{1}{c}{\textbf{Models}} & \multicolumn{1}{c}{\textbf{F1}} &  \multicolumn{1}{c}{\textbf{HEQ-Q}} &  \multicolumn{1}{c}{\textbf{HEQ-D}} &
\multicolumn{1}{c}{\textbf{Train time}}\\ \hline
   \hline
    BiDAF++ & 51.8/50.2 & 45.3/43.3 & 2.0/2.2& - \\
   BiDAF++ w/2Ctx & 60.6/60.1 & 55.7/54.8 & 5.3/4.0 & -\\ 
   \hline
    BERT + PHA & 61.8/- & 57.5/- & 4.7/- & 7.2\\
    BERT + PHQA & 62.0/- & 57.5/- & 5.4/-& 7.9 \\
    BERT + HAE & 63.1/62.4 & 58.6/57.8 & 6.0/5.1&10.1 \\
    \hline
    \textbf{BERT-CoQAC} & \textbf{65.1/64.0} & \textbf{60.2/59.6} & \textbf{6.6/5.8} & \textbf{8.9} \\
    \hline
    FlowQA & -/64.1 & -/59.6 & -/5.8 & 56.8\\
     \hhline{t:=:=:=:=:=:t}
  
    \end{tabular}
    \end{center}
    \caption{The evaluation results based on val/test scores. The top section is the baseline methods, the middle section is BERT-HAE with different methods and the bottom section lists the best performing models.}
    \label{result}
\end{table}
\begin{table}[]
\begin{center}
\begin{tabular}{cccc}\hline
    \hline
   \multicolumn{4}{c}{\textbf{Evaluation with 11 history turns on QuAC}}\\ 
    \hline
    \textbf{History Turns} &\textbf{F1} &\textbf{HEQ-Q} &\textbf{HEQ-D}\\
    \hline
    1 & 61.57 & 57.58 & 4.7 \\
    2 & 63.04 & 58.9 & 6.0\\
    3 & 62.58 & 58.64 & 5.4\\
    4 & 62.46 & 58.04 & 5.4\\
    \textbf{5} & \textbf{63.4} & \textbf{58.86} & \textbf{6.3}\\
    6 & 62.73 & 58.39 & 5.8  \\
    7 & 62.94 & 58.89 & 6.2\\
    8 & 62.16 & 58.10 & 4.6 \\
    9 & 62.9 & 58.05 & 5.6 \\
    10 & 62.23 & 58.26 & 5.7 \\
    11 & 62.13 & 58.11 & 5.6\\
    \hline
    \hline
\end{tabular}
\end{center}
\caption{The evaluation results of BERT-CoQAC with the varying number of turns on QuAC dataset.}
\label{turns}
\end{table}

\section{Conclusion and Future Work}
This paper is an effort to introduce a BERT-based framework, BERT-CoQAC, for effective conversational question answering in context. The proposed framework first selects the relevant history turns using the context-query relevance and models the history conversation by adding the history questions along with the embeddings to generate better context of the conversation. To verify our hypothesis, we conduct a number of experiments to analyze the effectiveness of relevant conversational turns on the overall accuracy of the model using a real-world dataset. The results show the effectiveness of our model.

The history selection mechanism used in the model is efficient but comprises of a very basic strategy. This paper is a work in progress and we plan on improvising the history selection strategy as future work.

%
%
%
\bibliographystyle{unsrt}
\bibliography{mybibtex}

\end{document}